\documentclass[journal]{IEEEtran}
\makeatletter
\def\ps@headings{%
\def\@oddhead{\mbox{}\scriptsize\rightmark \hfil \thepage}%
\def\@evenhead{\scriptsize\thepage \hfil \leftmark\mbox{}}%
\def\@oddfoot{}%
\def\@evenfoot{}}
\makeatother

\usepackage{graphics}
\usepackage{amsmath,graphicx}
\usepackage[lofdepth,lotdepth]{subfig}
\usepackage{paralist,cite}
\usepackage{amssymb,color}
\usepackage{algorithm2e}
\usepackage{algorithmic}

\newcommand{\nix}[1]{}

\begin{document}
\title{{A Missing and Found Recognition System \\for Hajj and Umrah\\}}
\author{
\bf{Salah A. Aly }\\
\medskip
Center of Research Excellence in Hajj and Umrah (HajjCore),\\ College of Computers and Information Systems,\\ Umm Al-Qura University, Makkah, KSA\\  salahaly@uqu.edu.sa
}
\maketitle

\begin{abstract}
This  note describes an integrated recognition system for identifying missing and found objects as well as missing, dead, and found people during Hajj and Umrah seasons in the two Holy cities of Makkah and Madina in the Kingdom of Saudi Arabia. It is assumed that the total estimated number of pilgrims will reach 20 millions during the next decade. The ultimate goal of this system is to integrate facial recognition and object identification solutions into the Hajj and Umrah rituals. The missing and found computerized system is part of the CrowdSensing system for Hajj and Umrah crowd estimation, management and safety.\footnote{--------------------------------------------------------------------------------- \goodbreak   Thanks to HajjCoRE, Center of Research Excellence in Hajj~ and~ Umrah at~ UQU,  an agency for supporting this work.}

\end{abstract}
%%
%\begin{keywords}
%Face detection, face recognition, and facial database sets.
%\end{keywords}
%%

\section{Introduction}
\label{sec:intro}

According to the Royal Embassy of Saudi Arabia, more than 10 million pilgrims (foreign and local) arrived to the Holy city of Makkah to perform Hajj and Umrah in 2011, among them, five million pilgrims performed Umrah during Ramadan and three million pilgrims performed Hajj. The total estimated number is expected to reach 20 million pilgrims per year during the next decade. Due to the huge crowd that occurs in the two Holy cities of Makkah and Madinah, several questions arise such as: How to identify missing, dead, and found people, and how to collect and distribute missing and found objects?

 One of the most crucial steps in the CrowdSensing system  is to detect  individuals' faces in an image and run a  facial recognition algorithm through a database of registered pilgrims~\cite{huer2012}. Therefore, the proposed approach is more focused on developing a pattern detection algorithm that does not depend on the three-dimensional complex data of faces, but it depends on the general outer Silhouette that is almost shared between all faces. The system goal is to develop a computational model for face detection and recognition that is reasonably simple and acceptably accurate under various conditions of  facial expressions, and various background environments.

The collected data regarding missing or found objects and people will be available throughout all year, and can be accessed online immediately from a PC or mobile device. The online missing and found database will be supported by a smart search engine for various search criterias. Also, the system can be deployed for border protection against illegal immigrants.

In this note, we describe a CrowdSensing system with two folds~\cite{crowdsensing,mfhajj}: \begin{compactitem} \item Identifying missing, dead, and found people during  overcrowds using facial detection and recognition, \item Collecting missing and found objects using a web-portal solution. \end{compactitem}

%%%%%%%%%%%%%%%%%%%%%%%%%%%%%%%%%%%%%%%%%%%%%%%%%%%%%%%%%%%%%%%%%%%%%%%%%%%%%%%%
\section{A Web-portal System for Missing and Found Items}

\begin{figure}[t]
  \begin{center}
  \includegraphics[width=8.5cm,height=5.5cm]{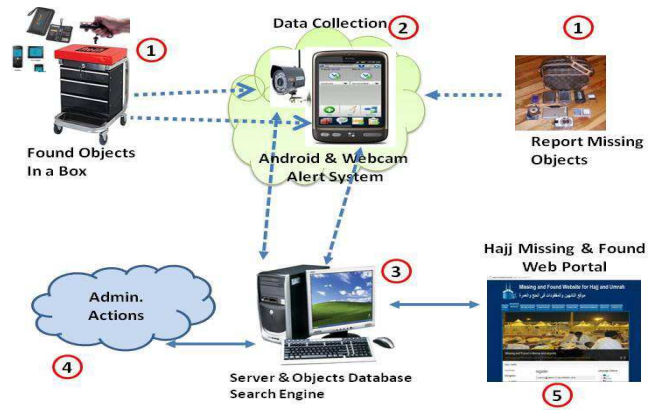}
  \caption{A proposed recognition system for missing and found objects during Hajj and Umrah seasons.}\label{fig:hajjmfobject}
  \end{center}
\end{figure}

During overcrowds in the two Holy cities during Hajj and Umrah, thousands of items are found and lost when moving around and getting-in sacred places.
The recognition system consists of the following parts, see Fig.~\ref{fig:hajjmfobject}:
\begin{enumerate}
  \item Boxes to collect found objects or personal items in Makkah and Madina. The Boxes can be placed in the main  intersection roads or near-by the two Holy Mosques.
  \item A touch screen system to be located in streets and public places to access the web-portals.
  \item A PC, IPAD, or mobile device with Android system to access the web-portal.
  \item  A main server to control and manage the system.
  \item A database engine to support all missing and found objects or personal items.
  \item Administrator and alerting system to take actions in cases of emergences.
\end{enumerate}

We  describe a MFHajj example as follows. Assume an object (watch or mobile device) is found. A description of this object along with a captured image will be entered into the web-portal via a mobile device supporting Android or by using one of the touch screens, which are located in the streets. Immediately, the information about this found object will be available online in the web-portal system, a search engine and object identification is supported in this system. The object owner can claim at any time, an evidence is given.

%%%%%%%%%%%%%%%%%%%%%%%%%%%%%%%%%%%%%%%%%%%%%%%%%%%%%%%%%%%%%%%%%%%%%%%%%%%%%%%%
\section{A Smart System for Recognizing Missing, Dead, and Found People}
\label{sec:systemdescription}

\begin{figure}[t]
  \begin{center}
  \includegraphics[width=8.3cm,height=6cm]{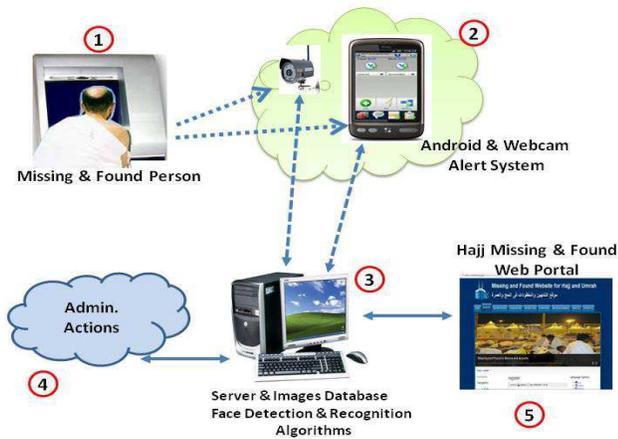}
  \caption{MFHajj portal interface website developed by the Crowdsensing.net team. The system consists of data collectors (mobiles, cameras, PCs), main server, search engine, and an alerting system. The system is used to recognize missing and found people during Hajj and Umrah seasons}\label{fig:hajjmfsyspeople}
  \end{center}
\end{figure}

During the overcrowds in Makkah and Madina, thousands of pilgrims are missing from their group members or can be found without any IDs. Furthermore, the majority of them do not know their location or the correct direction to their hotel rooms.
The system for identifying missing, dead, and found people consists of several parts as described below~\cite{mfhajj}, see Fig.~\ref{fig:hajjmfsyspeople}:
\begin{enumerate}
  \item A computer server to process the user's queries and to run the developed facial detection and recognition algorithms. The server will run a web-portal solution.
  \item A database search engine with all pilgrims' images~\cite{HUDA}. Such images can be easily captured at the KSA borders or in airport's entrance. Images of the local pilgrims can be collected when their Hajj permit cards are given.
  \item A mobile device supporting Android system that can send and receive online images, videos, and texts.
  \item A touch screen with webcam, which can be located in the main intersection roads or public places. Such a touch screen can be   similar to ATM machines.
        \item Administrator and alerting system to take actions in cases of emergences.

\end{enumerate}

The HUDA contains images taken during the 2011-2012 Hajj and  Umrah seasons of a large number of pilgrims (various races). It contains at least six images for each individual, in a varied range of poses, facial expressions (open and closed eyes, smiling/ not smiling), facial details (glasses/ no glasses), in  different lighting conditions, and against random backgrounds.
All images are in full-color JPG format, see Fig.~\ref{fig:fig1}.

For example, if a person is found alive or dead, his/her image can be taken by a mobile device supporting Android System. After that a query will be sent to the computer server to run the facial recognition algorithm and searching in the  pilgrims database, it will be conduced to watch his face, then extract his/her identity. Thereafter, a reply message immediately will be sent to the mobile device.
\subsection{Android and Software Portal Application}
\begin{figure}[t]
  \begin{center}
  \includegraphics[width=9cm,height=5.5cm]{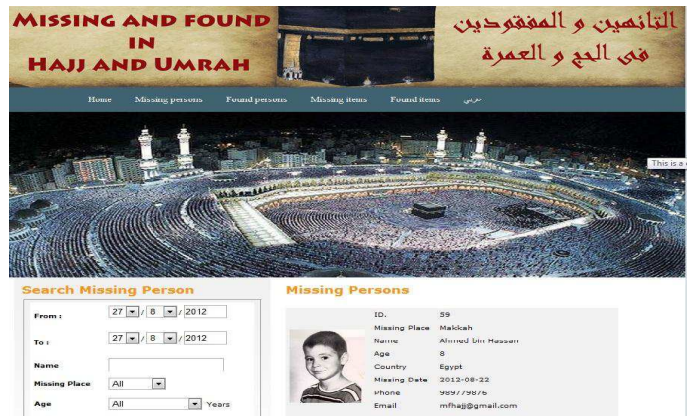}
  \caption{MFHajj portal interface website developed by Crowdsensing.net team. The system consists of data collectors (mobiles, cameras, PCs), main server, search engine, and an alerting system.}\label{fig:hajjmfsys}
  \end{center}
\end{figure}

We developed a software portal application to be installed in all mobile devices supporting Android System.  When the application runs in a user's mobile, an image will be taken and sent to a computer server for analysis. The server will consult the database system and will run facial recognition and detection algorithms.  Our developed software will use an image search engine as shown in Fig.~\ref{fig:hajjmfsys}. Also, an alerting system is used for an authority assistance. The software and portal applications are available in our project website shown in~\cite{crowdsensing}.

\subsection{Hajj and Umrah Face Recognition  Dataset}

The HUDA contains images taken during the 2011-2012 Hajj and Umrah seasons of a large number of pilgrims (varied races and appearances)~\cite{HUDA,crowdsensing}. It contains at least 3 images for each individual in a varied range of poses, facial expressions (open / closed eyes, smiling / not smiling) facial details (glasses / no glasses), in  different lighting conditions, and against random backgrounds. All the images are in full-color JPG format.

\begin{figure}[h]
\centering
    \includegraphics[width=0.4\textwidth]{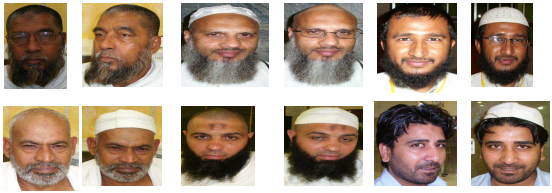}
    \caption{Samples  of Hajj and Umrah Dataset. One pilgrim image is stored in the MFHajj database and can be used in the search engine. Another different pilgrim image is taken to perform missing and found search process.}
    \label{fig:fig1}
\end{figure}

%%%%%%%%%%%%%%%%%%%%%%%%%%%%%%%%%%%%%%%%%%%%%%%%%%%%%%%%%%%%%%%%%%%%%%%%%%%%%%%%%%%%%%%5

\section{CrowdSensing System}
The CrowdSensing system is established to support
the existing efforts to manage the overcrowds and solve the missing
and found problem during Hajj and Umrah seasons in the two Holy cities of Makkah an Madina.
The goal of this CrowdSensing system is to deploy techniques
from Computer Vision and Image Processing to develop a portal
website for Hajj and Umrah missing and found people~\cite{crowdsensing}. The system
requires all pilgrims to register their personal data when they
plan to perform Hajj or Umrah. Such info can be collected easily when pilgrims cross the KSA borders.

One application of the proposed dataset is the CrowdSensing System, which consists of three main components, see Fig.~\ref{fig:hajjmfsys}:
 \begin{enumerate}
   \item A database of all individuals who arrive to the kingdom to perform the holy rituals. This database contains all their personal information along with a personal photo, and it can be updated via our web portal.
   \item  Advanced monitoring cameras scattered around the Grand Mosque in Makkah, airports, hospitals, and all areas of interest.
   \item  Our proposed face detection \& recognition algorithm is to be used for acquiring faces from images captured by the monitoring cameras and use them to identify missing and found individuals ~\cite{crowdsensing}.
 \end{enumerate}
There are several advantages of CrowdSensing system, some of which:
 \begin{enumerate}
   \item Estimating and counting crowds.
   \item Identifying missing, dead, and found people via face detection and recongition.
   \item Identifying missing and found items.
   \item Tracking pedestrians, and compute flow and density of crowds.
   \item Simulating crowds and movements.
   \item Recognizing empty spaces inside and outside El-Harram.
   \item Recognizing human event activities and event classifications.
   \item Collecting data and gathering information.
   \end{enumerate}

\section{Discussion and Conclusion}

In this note, we presented a computerized recognition system to solve the missing and found problem in Makkah and Madina during Hajj and Umrah seasons. A web-portal prototype is developed and has been tested in 2011 Hajj season~\cite{mfhajj}.

\bigskip

\section*{Acknowledgments}

This research is funded by a  grant number 11-nan1707-10 from the Long-Term National Plan for Science, Technology and Innovation (LT-NPSTI), the King Abdulaziz City for Science and Technology (KACST)- Kingdom of Saudi Arabia. We thank the Science and Technology Unit at Umm A-Qura University for their continued logistics support.

% -------------------------------------------------------------------------
\bibliographystyle{plain}

\begin{figure}[h]
\centering
    \includegraphics[width=8cm]{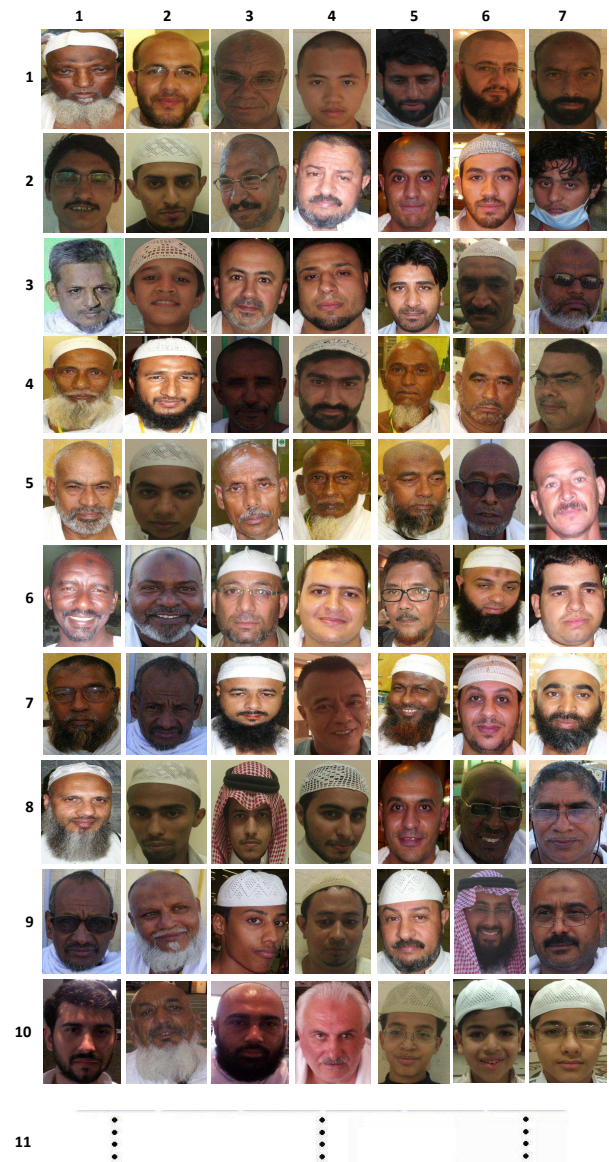}
    \caption{A sample  of Hajj and Umrah Dataset. }
    \label{fig:fig1}
\end{figure}

\end{document}